# Kernel Extreme Learning Machine Optimized by the Sparrow Search Algorithm for Hyperspectral Image Classification

**Zhixin Yan[1], Jiawei Huang[1], Kehua Xiang[1]**

[1] School of Computer Science, China West Normal University, Nanchong 637002, China

Corresponding author: Zhixin Yan (bravemr.yan@foxmail.com)

This work was financially supported by the School of Computer Science of China West Normal University.

**ABSTRACT** To improve the classification performance and generalization ability of the hyperspectral image classification algorithm, this paper uses Multi-Scale Total Variation (MSTV) to extract the spectral features, local binary pattern (LBP) to extract spatial features, and feature superposition to obtain the fused features of hyperspectral images. A new swarm intelligence optimization method with high convergence and strong global search capability, the Sparrow Search Algorithm (SSA), is used to optimize the kernel parameters and regularization coefficients of the Kernel Extreme Learning Machine (KELM). In summary, a multiscale fusion feature hyperspectral image classification method (MLS-KELM) is proposed in this paper. The Indian Pines, Pavia University and Houston 2013 datasets were selected to validate the classification performance of MLS-KELM, and the method was applied to ZY1-02D hyperspectral data. The experimental results show that MLS-KELM has better classification performance and generalization ability compared with other popular classification methods, and MLS-KELM shows its strong robustness in the small sample case.

## I. INTRODUCTION

In the past decades, remote sensing technology has developed at a high speed, and various spectral sensors with higher accuracy and resolution can capture many spectral ranges that are not visible to the human eye, such as near-infrared, mid-infrared, and far-infrared spectral intervals, making hyperspectral images not only contain textual information describing the space of features, but also rich spectral information responding to physical properties, which has led to the effective application of hyperspectral remote sensing technology in geological exploration, atmospheric monitoring, and environmental protection, etc. [1-3]. However, in the process of hyperspectral image application, many bottlenecks have emerged to limit its further development, among which the classification problem is one of the major problems to be solved in hyperspectral remote sensing [4].

At present, hyperspectral images face problems such as high cost, time-consuming, and difficulty of manual annotation of samples, which makes the number of annotated hyperspectral image samples limited, as well as problems such as the Hugh phenomenon which is very likely to occur in the process of hyperspectral image classification leading to poor classification accuracy. Many machine learning methods have been used to solve hyperspectral classification problems, such as support vector machine (SVM) [5, 6], random forest (RF) [7], etc. Although these methods achieve high classification accuracy, the high complexity of the computational process leads to a long-running time. In recent years, many deep learning methods have also been applied in the field of hyperspectral image classification [8,9], such as convolutional neural networks (CNN) [10-12], convolutional autoencoder, and convolutional neural networks (CAE-CNN) [13], and so on. However, most of these deep learning methods, which require manual construction of input parameters, have complex network structures and low efficiency [12]. In contrast, Broad Learning System (BLS) [14, 15] is an efficient incremental learning method without a deep network structure proposed by Chen et al. The network structure is simple and the classification speed is fast, but the method is only applicable to the case of smaller training data, and its effect is still not satisfactory for larger training data. In contrast, the Extreme Learning Machine (ELM) [16, 17] proposed by Huang et al. well avoids the disadvantages of the above methods, ELM is a hidden layer feedforward neural network, the input weights of the network and the bias values of the hidden layer nodes are determined randomly, which has the advantages of fast convergence and high accuracy. However, ELM still has shortcomings, as the hidden layer nodes of ELM are determined randomly, making it an uncertain network. The improved method Kernel Extreme Learning Machine [18, 19] solves this problem well by introducing kernel parameters and regularization coefficients based on inheriting the original advantages of ELM, thus it is no longer necessary to determine the hidden layer node size, which makes the network have faster classification speed and stronger generalization performance, but the kernel parameters and regularization coefficients of KELM still need to be determined manually[20].

In recent years, with the rise of swarm intelligence optimization methods, some swarm intelligence optimization methods have been used to solve the parameter optimization problem of KELM, such as the gray wolf optimization algorithm (GWO)[21, 22], whale optimization algorithm (WOA) [23, 24], particle swarm algorithm (PSO) [25], bat optimization algorithm (BA)[26] and other popular methods [27-29]. It is experimentally proven that although the above optimization methods can improve the performance of KELM to a certain extent, they still have problems such as poor convergence and easily fall into local optimal solutions. A new group intelligence optimization method with high convergence and strong global search capability, the sparrow search algorithm, was proposed by Xue et al. in 2020[30], which solves the above problems very well. Therefore, this paper uses the sparrow search method to optimize the above parameters, which saves the cost of manual tuning and improves the accuracy of the parameters.

Since hyperspectral images are characterized by a large number of bands, high spectral resolution, and large data redundancy, dimensionality reduction of hyperspectral data to extract effective spectral features is an indispensable step before the hyperspectral image classification. Many methods have been proposed to reduce spectral dimensionality for extracting spectral features and improving the discriminability of spectral information. Methods such as principal component analysis (PCA)[31-33], linear discriminant analysis (LDA)[34-36], and independent component analysis (ICA)[37-39], although these methods can effectively extract spectral features, they do not consider the spatial correlation between neighboring pixels in the process of feature extraction. In contrast, the Multi-Scale Total Variation method (MSTV)[40] proposed by Duan et al. achieves a more desirable feature extraction effect by taking advantage of the feature that hyperspectral images have multi-scale information, and noise reduction is applied to the spatial information of each scale separately while effectively extracting spectral features. And the method of feature processing by feature overlay [41] can make full use of the rich spectral information and spatial texture information contained in hyperspectral images to improve the subsequent classification accuracy in the feature extraction link. It has been proved that Local Binary Pattern [42] can effectively extract spatial texture features from hyperspectral data.

In summary, this paper proposes a hyperspectral classification method incorporating multiscale features based on the sparrow search method optimized kernel extreme learning machine. Firstly, the hyperspectral dataset is dimensionally reduced by MSTV to extract spectral features while noise reduction is applied to the adjacent spatial pixel points. After that and the reduced-dimensional dataset is extracted by LBP for spatial features. Feature superposition of spectral features and spatial features is performed to obtain fused features. The regularization coefficients and kernel parameters of the KELM are optimized by the SSA, and the optimal KELM classifier is obtained after training. Finally, the processed dataset is fed into the classifier to obtain the classification results.

The subsequent organization of this paper is as follows: Part II introduces the basic methods used in this paper. Part III introduces the high accuracy hyperspectral image classification algorithm proposed in this paper. Part IV sets up two experiments for verifying the rationality and feasibility of the MLS-KELM method. Part V discusses the results of this paper. Part VI is a summary of the whole paper.

## II. BASIC METHODS

### A. MSTV

MSTV is a feature extraction method for hyperspectral images based on multiscale full variance proposed by duan et al. in 2018. The MSTV method makes full use of the feature that hyperspectral images have multi-scale structural information, reduces the influence of texture information and noise by relative total variation (RTV), effectively reduces the differences within classes, and increases the variability between classes, making the extracted spectral features more discriminable.

The method is divided into three main steps. Firstly, the hyperspectral dataset is dimensionally reduced to reduce data redundancy and improve the efficiency of the whole method. Then, different parameters are constructed by RTV to extract the multi-scale structural features of the image, while reducing the effect of texture noise and fusing these spectral features of different dimensions. Finally, the fused multidimensional spectral features are scaled up using the kernel principal component analysis (KPCA) method to improve the separability of pixels belonging to different classes.

### B. LBP

LBP is used as an operator to describe the texture features of an image, which has the advantages of speed, grayscale invariance, and rotation invariance.

The original description of LBP is based on a pixel matrix of 3*3 size, and the pixel value of the middle pixel point is used as a threshold to compare with the surrounding pixel values. If the pixel value of a pixel point is greater than the threshold, the pixel point is marked as 1, and the opposite is marked as 0. After a round of comparison, the 3*3-pixel matrix generates an 8-bit binary LBP code, which is converted into decimal to represent. The LBP value is calculated by the following formula.

Formula (1) where denotes the central pixel point, denotes the grayscale value, is the surrounding pixel value, and is a symbolic function: $s(x) = \begin{cases} 1 & if\ x \geqslant 0 \\ 0 & else \end{cases}$

$$LBP(x_c, y_c) = \sum_{P=0}^{P-1} 2^P s(i_p - i_c) \quad (1)$$

### C. KELM

The Extreme Learning Machine, as a single hidden layer feedforward neural network, has a strong global search capability and fast classification speed. The network output of ELM can be represented by the following equation.

$$F(x) = h(x) \times \beta = H \times \beta = L \quad (2)$$

In formula (2), x is the input vector, h(x) and H are the hidden layer output matrix, β is the output weight, and L is the desired output. The KELM is an improved method based on ELM, which introduces the regularization coefficients and the unit matrix when solving the output weights of the hidden layer, which greatly enhances the stability of the learning network, and the output weights can be calculated by the following equation.

$$\beta = H^T \left( HH^T + \frac{I}{c} \right)^{-1} L \quad (3)$$

For the problems where the output matrix of the implicit layer is difficult to calculate and the amount of data is large, the KELM also introduces the kernel matrix, which can produce accurate classification results without solving the output matrix of the implicit layer, and the kernel matrix is calculated as follows.

$$\Omega_{ELM} = HH^T = h(x_i) \cdot h(x_j) = K(x_i, x_j) \quad (4)$$

In summary, the output matrix of KELM can be expressed as equation (5).

$$F(x) = h(x) H^T (I/C + HH^T)^{-1} Y$$
$$= \begin{pmatrix} K(x, x_1) \\ \vdots \\ K(x, x_n) \end{pmatrix}^T (1/C + \Omega_{ELM})^{-1} Y \quad (5)$$

### D. SSA

The sparrow search method is a new swarm intelligence optimization method proposed by Xue et al. in 2020, which has been applied many times to solve optimization problems for its advantages such as high convergence and strong global search capability. The SSA is based on the predatory and anti-predatory behavior of sparrows to derive the optimization method. In the predation and anti-predation behavior of sparrows, the sparrow population is divided into observers, joiners, and scouts. Among them, the observer is the most adaptable and has the strongest exploration ability, so it is responsible for detecting the location of prey, and once the prey is found, it will notify the rest of the sparrows to come and hunt. When the joiner receives the predation signal from the observer, it will go forward to hunt. The scouts are responsible for detecting invaders and sending warning signals to the sparrow population when danger is detected.

### III. Hyperspectral Image Classification Methods

#### A. Introduction of MLS-KELM

The MSTV method can reduce the dimensionality of hyperspectral data to reduce data redundancy and improve the discriminability of spectral features. Compared with other dimensionality reduction methods, such as PCA, ICA, and other mainstream dimensionality reduction algorithms, MSTV reduces the spatial correlation of neighboring pixels using the RTV algorithm while giving dimensionality reduction to hyperspectral data. It enables to reduce of the adverse effect of texture noise and the feature extraction effect is improved. Hyperspectral images contain rich spatial texture information in addition to spectral information. LBP is an efficient operator for describing image texture features, which can be used to extract spatial features of hyperspectral data with the advantages of gray invariance, rotation invariance, and fast. KELM is an improved method of ELM, which introduces kernel parameters and regularization coefficients compared with the traditional feedforward neural network model. The SSA method is a new type of swarm intelligence optimization method, which has the advantages of fewer parameters, fast convergence, and strong global searchability.

In summary, this paper proposes a hyperspectral image classification method based on an optimized KELM with fused multiscale features--MLS-KELM. Firstly, the spectral data dimensionality is reduced by the MSTV method, the spectral features are extracted, and the spatial texture information is done with noise reduction. Then the spatial features of hyperspectral data are extracted using LBP. The spectral features and spatial features are fused. The KELM optimized by SSA is used as an image classifier. Input validation set to the classifier to get the classification results.

## B. The flow of the MLS-KELM

The model diagram of the MLS-KELM method is shown in Figure.1.

The method steps are as follows.

**Step1: Extract spectral features and perform noise reduction on spatial information**

Firstly, the data set with the number of bands M is divided into K groups, where K is the number of dimensions to be dimensioned down. In this paper, K is taken as 20, where the kth subset can be obtained from (6).

$$P^k = \begin{cases} (I_{(k-1)\lfloor M/K \rfloor +1}, \cdots, I_{(k\lfloor M/K \rfloor)}), & if\ k\lfloor M/K \rfloor \leq M \\ (I_{(K-\lfloor M/K \rfloor +1)}, \cdots, I_M), & if\ k\lfloor M/K \rfloor > M \end{cases} \quad (6)$$

Then calculate the data of each band after dimensionality reduction. The reduced dimensional dataset Q is obtained.

$$Q^k = \frac{\sum_{n=1}^{N_k} P_n^k}{N_k} \quad (7)$$

Set the RTV model with different parameters and input the reduced-dimensional dataset to obtain the stacked multi-scale structural features:

$$F_l = RTV(Q, \lambda_l, \sigma_l),\ l \epsilon \{1, 2, \cdots, L\} \quad (8)$$

The multi-scale structural features are fused by kernel principal component analysis. where N is the number of principal components to be retained, and P is the dataset with the spectral features extracted and processed by noise reduction.

$$P = KPCA(F, N) \quad (9)$$

**Step2: Extraction of spatial features**

The spatial features of hyperspectral data are extracted by LBP, an efficient operator for describing texture features, which is defined in its original form in a 3*3-pixel matrix. The middle pixel value is compared with the surrounding pixel values in clockwise order, and the local binary pattern of each central pixel point is traversed in turn by equation (1). The eigenvalues are converted into decimal form, which is the spatial feature of this hyperspectral image.

**Step3: Feature fusion**

To ensure that the scale of the feature values is consistent, the extracted spectral features and spatial features are normalized separately, i.e., all the feature values are mapped to between 0 and 1. The spectral features and spatial features are directly fused by using the feature superposition method to form a long feature vector to obtain the fused features.

**Step4: Optimizing the KELM classifier**

In this paper, the activation function of KELM is chosen as RBF for setting the kernel parameter K. When calculating the output matrix H of the hidden layer, the introduction of the regularization coefficient can make the output matrix more stable. Therefore, the kernel parameters and regularization coefficients largely affect the performance of KELM. Consequently, SSA is used to optimize the above parameters in this paper. First, the hyperspectral dataset is divided into the training set and the validation set, and the fitness value is chosen as the mean square error (MSE) of the training set. When the SSA method reaches the maximum number of iterations, the optimal solution of the output is assigned to the kernel parameter and regularization coefficient of KELM, which is the optimized KELM classifier. The optimization process is shown in Figure.2.

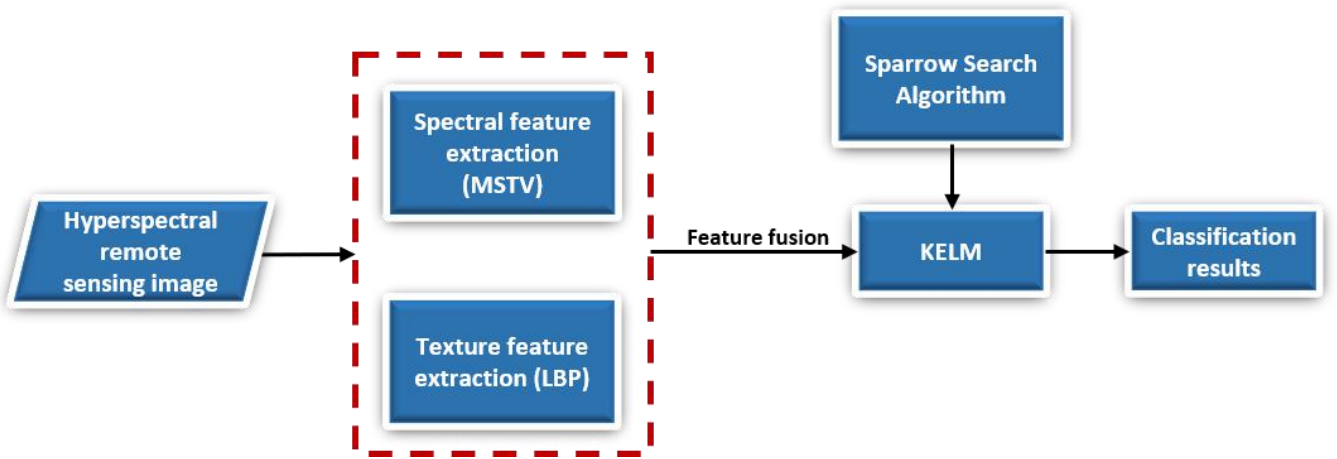

**FIGURE.1 Model diagram of the MLS-KELM**

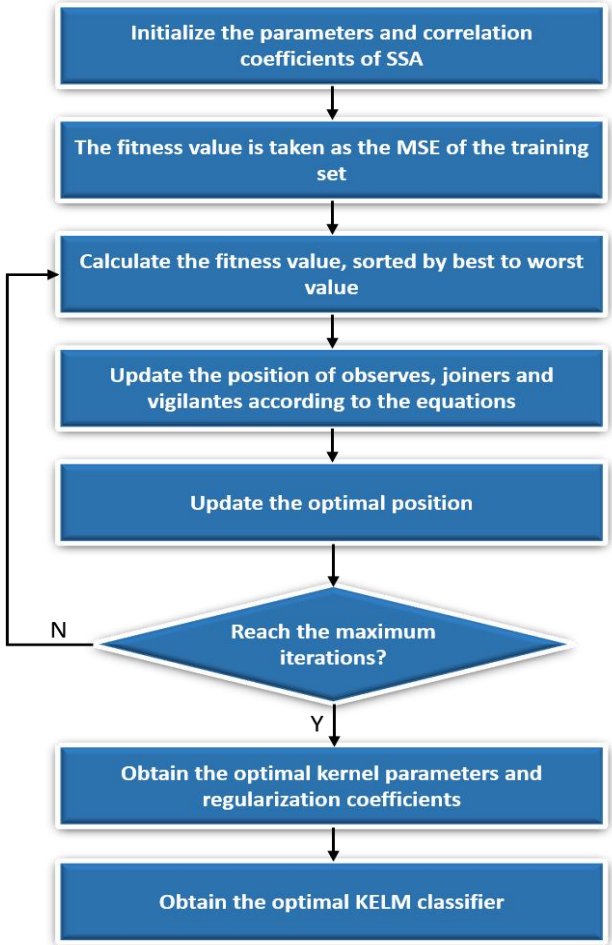

**FIGURE.2** Steps of parameter optimization

The steps of parameter optimization are described as follows.
1) Initialize the population size of SSA and define the correlation coefficient.
2) Calculate the fitness value based on the MSE of the training set. The best and worst individual positions of the population are obtained by ranking the fitness degrees.
3) Update the observer positions according to Equation (10).

$$D_{c,e}^{t+1} = \begin{cases} D_{c,e}^t \cdot \exp\left(-\dfrac{c}{\alpha iter_{\max}}\right), & R_2 < ST \\ D_{c,e}^t + QK, & R_2 \geqslant ST \end{cases} \quad (10)$$

In equation (10), $t$ is the current number of iterations, $aiter_{max}$ is the maximum number of iterations, D denotes the position information of the $c_{th}$ sparrow individual in the $e_{th}$ dimension during the $t_{th}$ iteration, α is a random number between 0 and 1, $R_2$ is a warning value whose range is between 0 and 1, $ST$ is a safety value whose range is between 0 and 0.5, and $Q$ is a random number conforming to a positive-terrestrial distribution.

4) Update the joiner positions according to Equation (11).

$$D_{c,e}^{t+1} = \begin{cases} Q_{\exp}\left(\dfrac{D_{worst}^t - D_{c,e}^t}{c^2}\right), & c > \dfrac{n}{2} \\ D_F^{t+1} + |D_{c,e}^t - D_F^{t+1}| \cdot A^+ \cdot K, & otherwise \end{cases} \quad (11)$$

Equation (11), $D_F^{t+1}$ denotes the current sparrow's best position; $D_{worst}^t$ denotes the current worst position; A is a 1*dimensional matrix whose values are randomly assigned as 1 or -1; $A^+$ is a pseudo-inverse matrix whose values are calculated by $A^+ = A^T(AA^T)^T$.
5) Update the location of the vigilantes according to Equation (12).

$$D_{c,e}^{t+1} = \begin{cases} D_{best}^t + V|D_{c,e}^t - D_{best}^t|, & f_c > f_g \\ D_{c,e}^t + O\left[\dfrac{|D_{c,e}^t - D_{worst}^t|}{(f_c - f_w) + \delta}\right], & f_c = f_g \end{cases} \quad (12)$$

In equation (12), $D_{best}$ denotes the global optimal position, V denotes the step control parameter, whose value takes 0 or 1, and $O$ is a random number distributed in [-1,1], which denotes the control parameter of the sparrow's movement step. $f_c$ denotes the current adaptation value of the sparrow, $f_g$ and $f_w$ denotes the current best adaptation value and the worst adaptation value, respectively. $\delta$ is a minimal value constant, which is used to avoid the case that the denominator is 0.
6) The current latest position is obtained by formulas (10)-(17), and if the latest position is better than the optimal position, the value of the optimal position is replaced with the current latest position.
7) Determine whether the maximum number of iterations is reached, if yes, the iteration is finished, otherwise go back to 3).
8) Output the optimal result and assign it to the kernel parameter and regularization coefficient to obtain the optimal KELM classifier.

**Step5: Deriving classification results**
The validation set is fed into the optimal KELM classifier to derive the predicted classification labels. The OA, AA, and Kappa coefficients of this classification result are calculated to obtain the image classification result.

## IV. EXPERIMENTAL

### A. Experimental Environment Description
All experiments in this paper are done on a device with Intel(R) Core(TM) i7-10875H CPU and 16GB RAM. The programming language is Matlab, and the test software version is Matlab 2020b.

### B. Dataset introduction

To verify the optimization performance of SSA and the classification performance of the MLS-KELM method, three publicly available hyperspectral remote sensing image datasets are selected for validating in this paper, namely Indian Pines dataset, Houston 2013 dataset, and Pavia University dataset.

Indian Pines dataset was acquired by AVIRIS sensor from an Indian pine forest in northwestern USA, which contains a total of 10249 pixels of features, including 16 types of agricultural and forestry features such as corn, grass, wheat, and trees. The Houston 2013 dataset was collected by the CASI-1500 sensor from the urban area around the University of Houston, USA. It contains a total of 15,029 pixels of features, including water, trees, dirt, roads, highways, and other 15 categories of urban features. The Pavia University dataset was collected by the ROSIS sensor from the urban area around the University of Pavia, Italy. These pixels contain a total of 9 types of urban features, including trees, asphalt roads, bricks, pastures, etc.

Table I shows the basic information of the three datasets. The three hyperspectral datasets after preprocessing were divided into validation and training sets, and two cases of 10% training set and 5% training set were set. Table II shows the results of training set partitioning for the three datasets.

TABLE I.
BASIC INFORMATION OF THE THREE DATASETS

| Data | Indian Pines | Pavia University | Houston 2013 |
|---|---|---|---|
| Collection site | Indiana, USA | Northern Italy | Houston, USA |
| Acquisition device | AVIRIS | ROSIS | CASI-1500 |
| Spectral range (μm) | 0.4~2.5 | 0.43~0.86 | 0.38~1.05 |
| Data size(pixel) | 145×145 | 610x340 | 349×1905 |
| Spatial resolution(m) | 20 | 1.3 | 2.5 |
| Number of bands | 220 | 115 | 144 |
| Number of bands after denoising | 200 | 103 | 144 |
| Sample size | 10249 | 42776 | 15029 |
| Number of categories | 16 | 9 | 15 |

TABLE II.
TRAINING SET PARTITIONING RESULTS FOR THE THREE DATASETS

| Percentage of training sets | Data | Indian Pines | Pavia University | Houston 2013 |
|---|---|---|---|---|
| 10% | Number of training sets | 1025 | 4278 | 1503 |
| | Number of test sets | 9224 | 38498 | 13526 |
| 5% | Number of training sets | 512 | 2139 | 751 |
| | Number of test sets | 9224 | 40637 | 14278 |
| Total number of samples | | 10249 | 42776 | 15029 |

### C. Experimental description

To verify the convergence and global search ability of SSA in processing hyperspectral data and the actual classification effect of the MLS-KELM method. In this paper, two sets of experiments are set up: in the first set of experiments, the pre-processed hyperspectral dataset is taken as the input sample, which is iteratively optimized by using the SSA and each popular swarm intelligence optimization method, and the mean, variance, optimal value and worst value of each method are counted. The convergence graphs are drawn to compare the convergence and the search performance of each optimization method. Among the mainstream optimization methods, we select the gray wolf optimization algorithm (GWO), whale optimization algorithm (WOA), particle swarm optimization algorithm (PSO), and bat optimization algorithm (BA). In the second group of experiments, the MLS-KELM method proposed in this paper is compared with the traditional classification method, and the classification effect graph and accuracy comparison graph are made to verify the actual classification effect of MLS-KELM. The evaluation indexes are OA, AA, and Kappa coefficient. The popular classification methods are selected as PCA-CNN, CAE-CNN, BLS, SVM, and KELM.

### D. Experiment 1: Verifying the optimization performance of SSA

The three hyperspectral datasets after pre-processing were set for two cases of 10% training set and 5% training set. The training set is used as the input sample of the optimization method, and the sparrow search method is compared with each popular swarm intelligence optimization method, and the best value (best), worst value (worst), mean value (mean), and variance (std) of the solutions of each method are counted, where the mean value can be used to measure the optimization accuracy of the method, and the accuracy rank (rank) of each method is obtained according to the mean value, and the iterative The convergence curve of the optimization is drawn to observe and verify its convergence. Among the popular swarm intelligence optimization methods we select the GWO, WOA, PSO, and BA. The fitness function is selected as the MSE of the training set, the number of populations is set to 30, and the maximum number of iterations is set to 20.

#### 1) ACCURACY ANALYSIS

For processing hyperspectral data, each optimization algorithm achieves good optimization results. For Indian Pines, the average accuracy of the GWO is slightly higher than that of the WOA and much better than that of the BA and PSO, while the average accuracy of the SSA is better. For Pavia University, the average precision of the WOA is slightly higher than that of the GWO and much better than that of the BA and PSO, while the average precision of the SSA is optimal. For Houston 2013, the average optimized accuracy of the WOA is slightly higher than that of the GWO and much better than that of the BA and PSO, and the average optimized accuracy of the SSA is still the highest.

In summary, for processing hyperspectral data, both GWO and WOA can achieve good optimization accuracy, BA and PSO are relatively poor, and SSA is the best performer in terms of optimization. Analysis of the optimization results shows that the mean value of the solution of SSA is optimal under the three data sets tested, indicating that it indicates that SSA has the highest quality of optimization accuracy and solution. Its optimal values are all better than other algorithms, indicating that it always finds the optimal solution, which shows that SSA has a stronger global search capability.

TABLE III.
OPTIMIZATION RESULTS OF EACH HYPERSPECTRAL DATASET

| Dataset | Percentage of training set | Statistical values | SSA | GWO | WOA | PSO | BA |
|---|---|---|---|---|---|---|---|
| Indian Pines | 10% | mean | 1.24E-02 | 1.30E-02 | 1.32E-02 | 1.95E-02 | 1.59E-02 |
| | | std | 1.43E-07 | 2.66E-05 | 3.89E-05 | 2.17E-01 | 3.38E-04 |
| | | worst | 1.38E-02 | 1.38E-02 | 1.64E-02 | 1.97E-02 | 1.66E-02 |
| | | best | 1.21E-02 | 1.29E-02 | 1.29E-02 | 1.93E-02 | 1.57E-02 |
| | | rank | 1 | 2 | 3 | 5 | 4 |
| | 5% | mean | 4.71E-02 | 4.81E-02 | 4.82E-02 | 5.36E-02 | 5.13E-02 |
| | | std | 1.45E-06 | 2.08E-05 | 2.95E-05 | 2.17E-04 | 4.41E-04 |
| | | worst | 5.02E-02 | 4.98E-02 | 5.11E-02 | 5.44E-02 | 5.30E-02 |
| | | best | 4.68E-02 | 4.78E-02 | 4.78E-02 | 5.31E-02 | 5.05E-02 |
| | | rank | 1 | 2 | 3 | 5 | 4 |
| Pavia University | 10% | mean | 2.66E-02 | 2.71E-02 | 2.69E-02 | 3.05E-02 | 2.83E-02 |
| | | std | 3.14E-06 | 6.63E-06 | 1.58E-05 | 8.73E-03 | 6.13E-04 |
| | | worst | 2.77E-02 | 2.99E-02 | 2.73E-02 | 3.19E-02 | 2.93E-02 |
| | | best | 2.65E-02 | 2.68E-02 | 2.68E-02 | 2.98E-02 | 2.78E-02 |
| | | rank | 1 | 3 | 2 | 5 | 4 |
| | 5% | mean | 5.95E-02 | 6.02E-02 | 6.00E-02 | 6.84E-02 | 6.35E-02 |
| | | std | 7.84E-06 | 1.01E-05 | 5.90E-04 | 2.31E-03 | 1.77E-03 |
| | | worst | 6.25E-02 | 6.43E-02 | 6.25E-02 | 7.39E-02 | 6.68E-02 |
| | | best | 5.92E-02 | 5.98E-02 | 5.98E-02 | 6.67E-02 | 6.14E-02 |
| | | rank | 1 | 3 | 2 | 5 | 4 |
| Houston 2013 | 10% | mean | 7.29E-03 | 7.51E-03 | 7.43E-03 | 9.12E-02 | 7.92E-02 |
| | | std | 6.50E-08 | 4.50E-06 | 7.48E-04 | 3.06E-04 | 2.38E-04 |
| | | worst | 7.90E-03 | 1.03E-02 | 1.07E-02 | 1.03E-02 | 8.86E-03 |
| | | best | 6.96E-03 | 8.43E-03 | 7.19E-03 | 8.98E-03 | 7.83E-03 |
| | | rank | 1 | 3 | 2 | 5 | 4 |
| | 5% | mean | 2.45E-03 | 2.54E-03 | 2.48E-03 | 3.18E-02 | 2.56E-02 |
| | | std | 2.55E-07 | 3.88E-04 | 2.05E-03 | 1.23E-02 | 5.23E-02 |
| | | worst | 3.13E-02 | 3.17E-02 | 3.36E-02 | 3.33E-02 | 3.96E-02 |
| | | best | 2.36E-02 | 2.42E-02 | 2.42E-02 | 3.03E-02 | 2.52E-02 |
| | | rank | 1 | 3 | 2 | 5 | 4 |

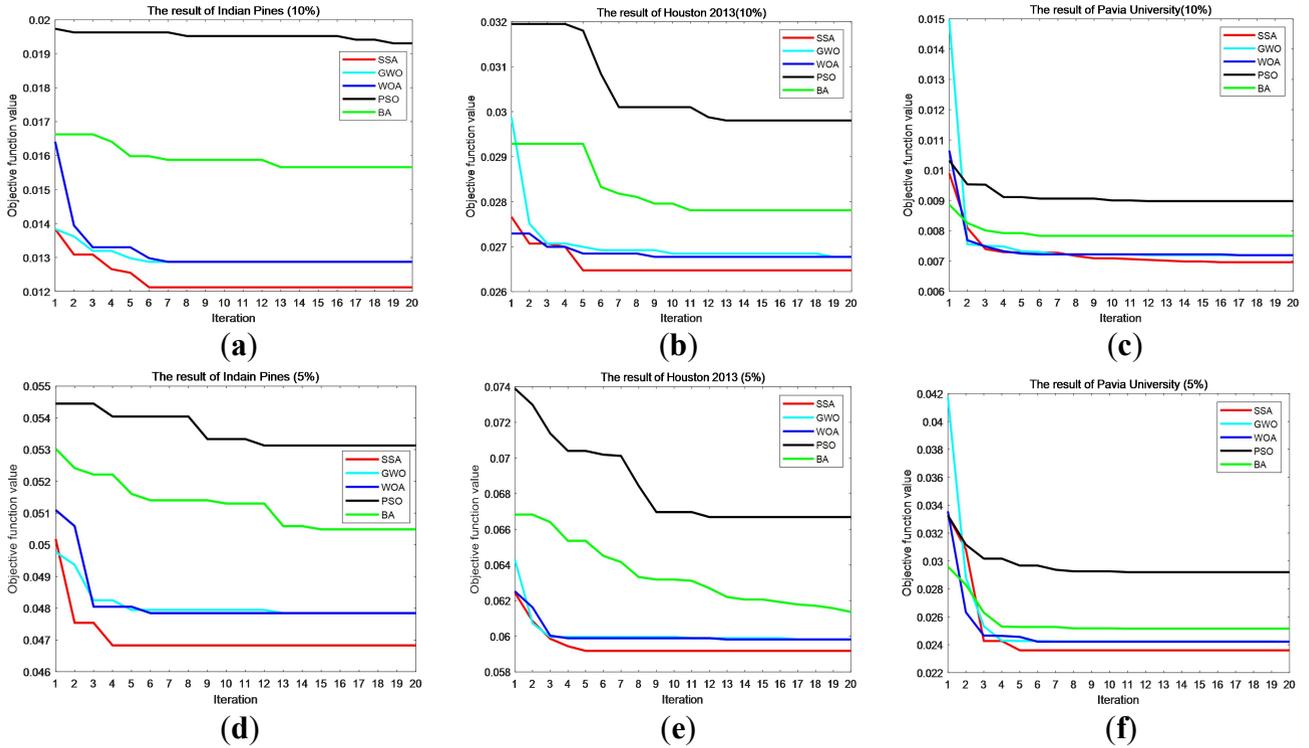

**FIGURE.3** Convergence curve of each group intelligent optimization method. (a) Indian Pines (10%); (b) Houston 2013 (10%); (c) Pavia University (10%); (d) Indian Pines (5%); (e) Houston 2013 (5%); (f) Pavia University (5%).

### 2) CONVERGENCE ANALYSIS

The smaller standard deviation represents the better stability of the algorithm. Analysis of Table III shows that for Indian Pines, the stability of GWO and WOA is better than that of PSO and BA, while SSA has the smallest standard deviation and the best stability. For Pavia University, the standard deviation of GWO is significantly smaller than that of WOA, PSO, and BA, while the standard deviation of SSA is smaller than that of GWO, indicating that SSA has the most stable performance. For Houston 2013, the standard deviation of GWO is significantly smaller than that of WOA, PSO, and BA, which shows good stability, while the standard deviation of SSA is smaller, so SSA has the strongest stability.

In summary, it can be seen that for processing hyperspectral datasets, GWO has good stability, while SSA is relatively better.

### 3) CONVERGENCE ANALYSIS

In Figure 3, the convergence curves of each method are plotted separately for different datasets and training set percentages. The convergence curves show that SSA always obtains higher classification accuracy for the same number of iterations. SSA can find the global optimal solution with fewer iterations and maintain its stable fast convergence for different datasets and training set percentages. Compared with the rest of the swarm intelligence optimization methods, SSA has better convergence and stability.

In summary, for processing hyperspectral remote sensing data, SSA still maintains its advantages of high convergence, strong global searchability, and high accuracy. It can always find the global optimal solution in a relatively short time. It can be seen that SSA is more competitive than other advanced swarm intelligence optimization algorithms in the field of hyperspectral image classification.

### E. Experiment 2: MLS-KELM performance verification

To verify the classification accuracy and generalization performance of the MLS-KELM method, the method is compared with the popular classification method, and the classification effect of each method is tested under three datasets, and the classification effect graph and accuracy comparison graph are made to verify the actual classification effect of MLS-KELM by analyzing the experimental results. The evaluation indexes are OA, AA, and Kappa coefficient. The popular classification methods are selected as PCA-CNN [43], CAE-CNN, BLS, SVM, and KELM. TABLE IV, TABLE V, and TABLE VI show the classification results of each method in Indian Pines, Pavia University, Houston 2013 dataset respectively.

1) INDIAN PINES DATASET

TABLE IV.
CLASSIFICATION RESULTS OF INDIAN PINES DATASET

| Percentage of training set (%) | Evaluation metrics | PCA-CNN | CAE-CNN | BLS | SVM | KELM | MLS-KELM |
|---|---|---|---|---|---|---|---|
| 10% | OA (%) | 90.43% | 91.79% | 97.70% | 98.28% | 98.32% | 98.78% |
|  | AA (%) | 84.56% | 86.44% | 94.19% | 98.00% | 97.54% | 98.50% |
|  | Kappa (%) | 89.11% | 90.66% | 97.38% | 98.04% | 98.08% | 98.61% |
|  | Tra-time(s) | 331.56 | 345.01 | 91.65 | 153.55 | 14.60 | 24.33 |
| 5% | OA (%) | 86.51% | 90.44% | 94.20% | 95.04% | 95.01% | 95.22% |
|  | AA (%) | 78.34% | 84.60% | 92.73% | 92.74% | 93.61% | 93.99% |
|  | Kappa (%) | 84.61% | 89.13% | 93.40% | 94.36% | 94.32% | 94.56% |
|  | Tra-time(s) | 211.09 | 219.32 | 85.21 | 118.59 | 11.74 | 18.22 |

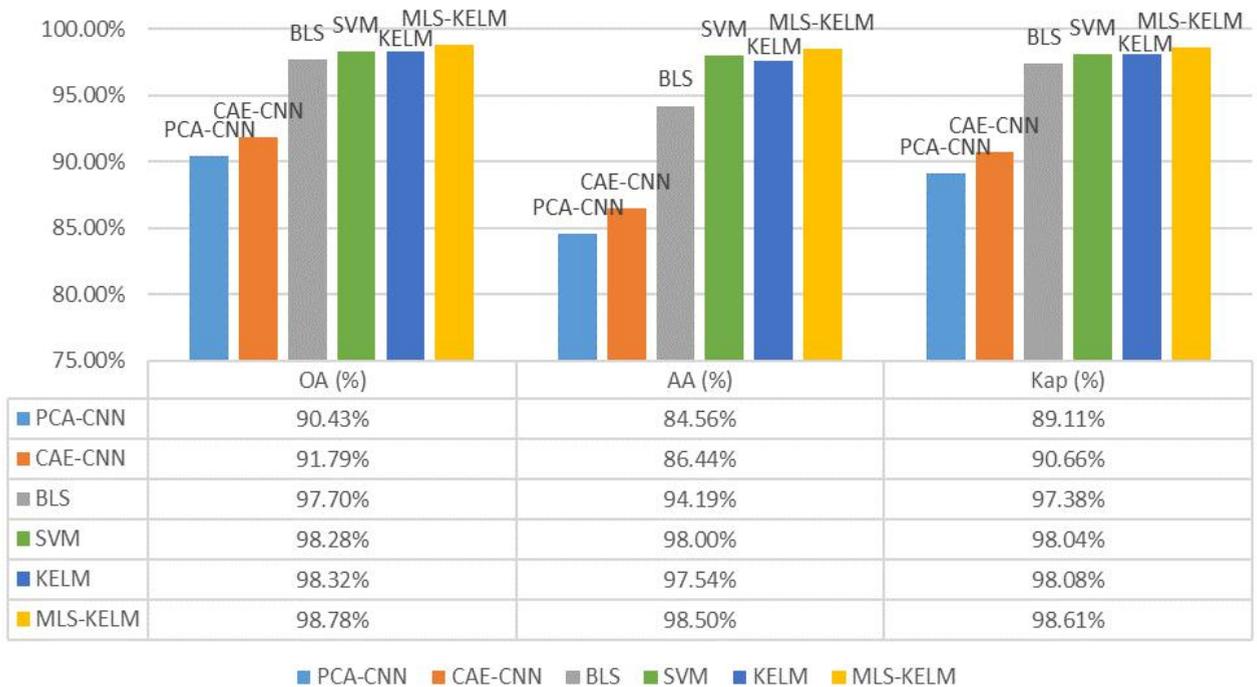

FIGURE.4 Classification comparison results on Indian Pines dataset (10%)

The analysis of the classification effect of each method under Indian Pines dataset by experimental results is as follows: under the 10% training set, the OA, AA, and Kappa coefficients of KELM reach 98.32%, 97.54%, and 98.08% respectively, which is significantly higher than PCA-CNN, CAE-CNN, BLS, and SVM. MLS-KELM achieves 98.78%, 98.50%, and 98.61% for OA, AA, and Kappa, respectively, on which its OA, AA, and Kappa coefficients are improved by 0.46%, 0.96%, and 0.53%, respectively. In contrast, the OA, AA, and Kappa coefficients of the MLS-KELM are improved by 3.48%, 6.35%, and 3.96% on average with 10% training data compared with the traditional PCA-CNN, CAE-CNN, BLS, SVM, and KELM. Significantly better than PCA-CNN, CAE-CNN, BLS, SVM, and KELM methods. With a 5% training set, MLS-KELM improves the OA, AA, and Kappa coefficients by 0.21%, 0.38%, and 0.24%, respectively, compared to the KELM method. The experimental results show that MLS-KELM has high classification accuracy on Indian Pines and maintains a high classification performance in the case of small samples, indicating its strong robustness.

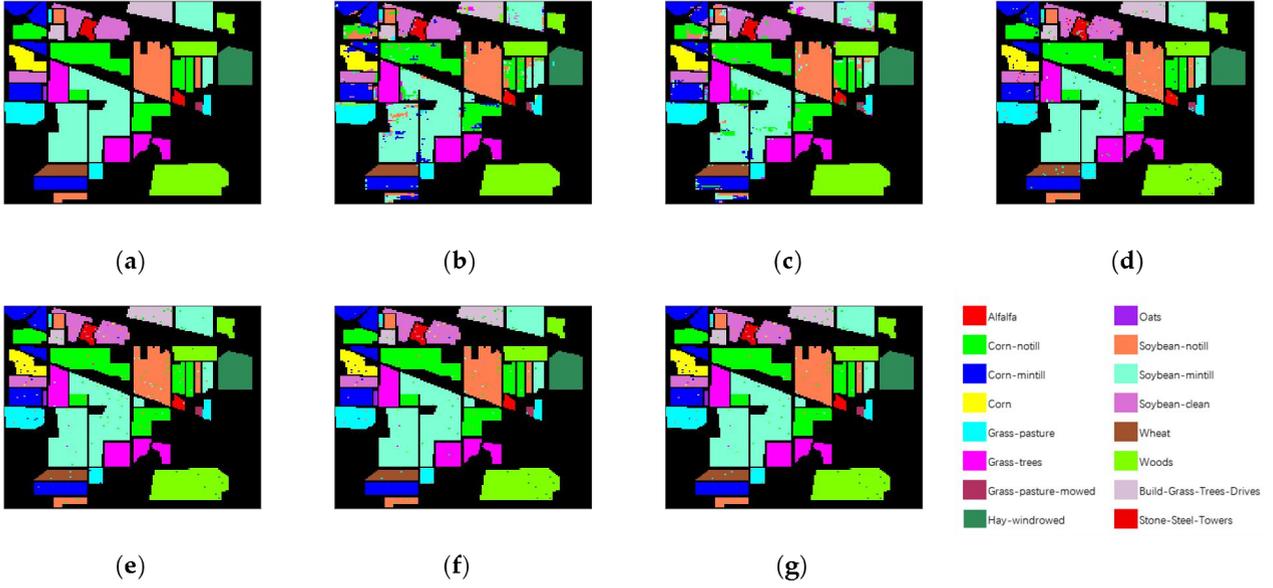

**FIGURE.5** Classification effect of Indian Pines dataset (10%). (a) Original graph; (b) PCA-CNN; (c) CAE-CNN; (d) BLS; (e) SVM; (f) KELM; (g) MLS-KELM.

2) PAVIA UNIVERSITY DATASET

TABLE V.
CLASSIFICATION RESULTS OF PAVIA UNIVERSITY DATASET

| Percentage of training set (%) | Evaluation metrics | PCA-CNN | CAE-CNN | SVM | BLS | KELM | MLS-KELM |
|---|---|---|---|---|---|---|---|
| 10% | OA (%) | 98.44% | 98.59% | 99.26% | 77.19% | 99.04% | 99.28% |
| | AA (%) | 98.61% | 98.94% | 99.51% | 84.34% | 99.39% | 99.54% |
| | Kappa (%) | 97.84% | 98.07% | 98.99% | 71.10% | 98.67% | 99.01% |
| | Tra-time(s) | 722.36 | 745.01 | 431.65 | 19.55 | 14.6 | 124.33 |
| 5% | OA (%) | 97.10% | 97.48% | 97.07% | 77.31% | 97.31% | 97.58% |
| | AA (%) | 97.34% | 97.64% | 96.46% | 84.87% | 97.80% | 98.09% |
| | Kappa (%) | 96.09% | 96.60% | 96.07% | 71.60% | 96.38% | 96.74% |
| | Tra-time(s) | 446.57 | 379.32 | 385.21 | 18.59 | 10.73 | 68.22 |

The analysis of the classification effects of each method under the Pavia University dataset by experimental results is as follows: with a 10% training set, the OA, AA, and Kappa coefficients of KELM reach 99.04%, 99.39%, and 98.67%, respectively, and the classification effects are significantly higher than those of PCA-CNN, CAE-CNN, BLS. And the OA, AA, and Kappa of MLS-KELM reached 99.28%, 99.54%, and 99.01%, respectively, on which its OA, AA, and Kappa coefficients improved by 0.25%, 0.16%, and 0.34%, respectively, with the highest classification accuracy among the five groups of methods. The MLS-KELM method, on the other hand, improves its OA, AA, and Kappa coefficients by 4.78%, 3.38%, and 6.08% on average with 10% training data compared to the popular PCA-CNN, CAE-CNN, BLS, SVM, and KELM methods. Significantly better than PCA-CNN, CAE-CNN, BLS, SVM, and KELM methods. With a 5% training set, MLS-KELM improves the OA, AA, and Kappa coefficients by 0.28%, 0.30%, and 0.38%, respectively, compared to the KELM method, and 0.03%, 0.14%, and 0.04%, respectively, relative to the 10% training set. The experimental results show that MLS-KELM has high classification accuracy on Pavia University and maintains a high classification performance in the case of small samples, indicating its strong robustness.

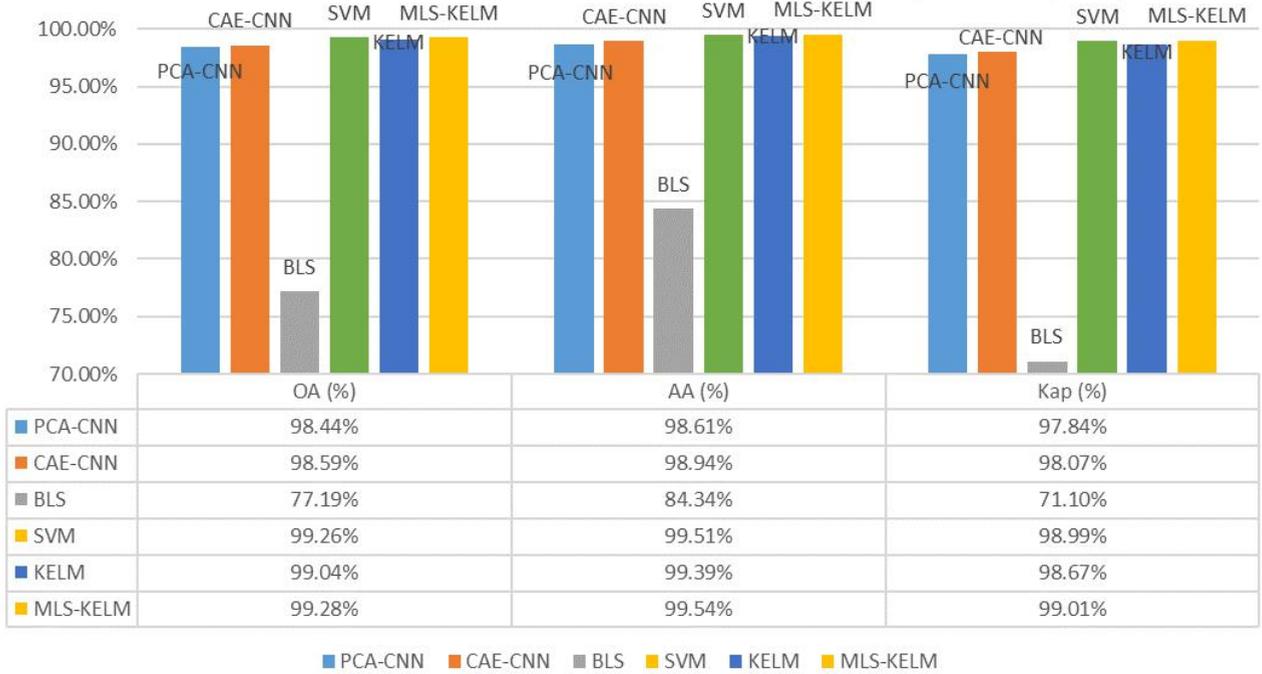

**FIGURE.6** Classification comparison results on Pavia University dataset

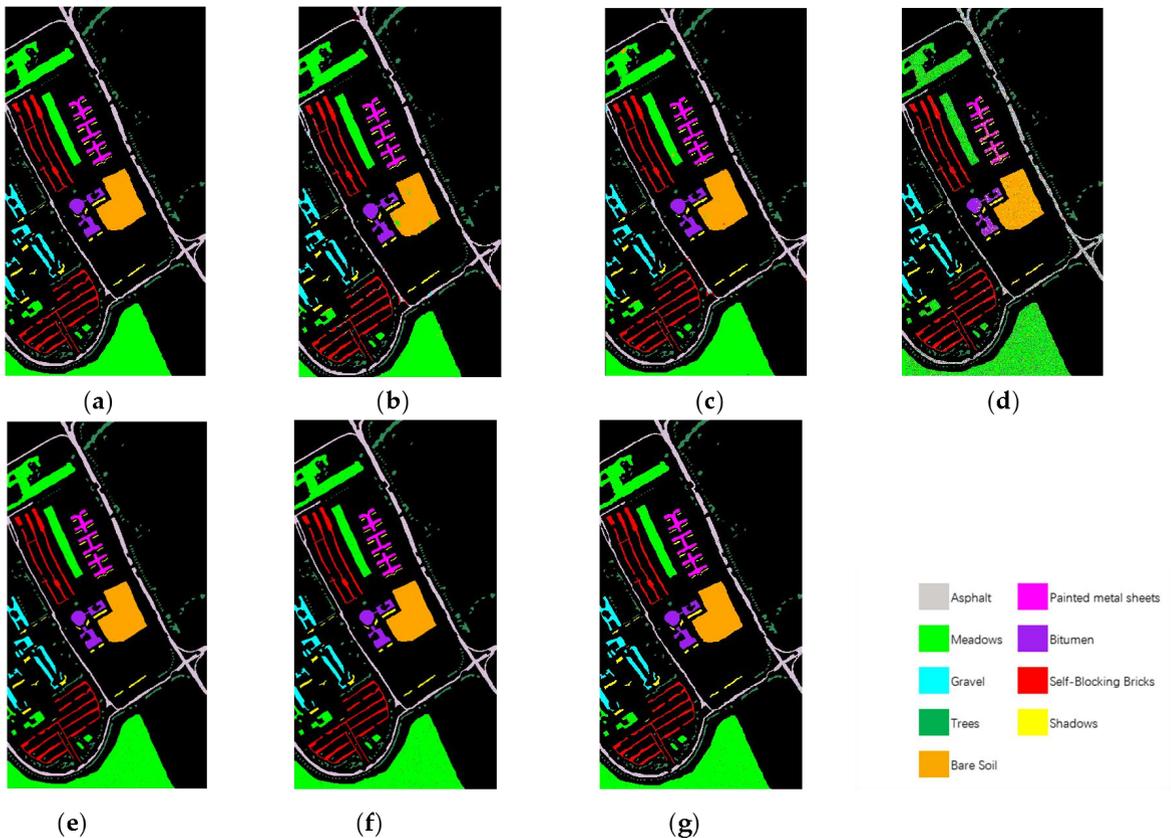

**FIGURE.7** Classification effect of Pavia University dataset (10%). (a) Original graph; (b) PCA-CNN; (c) CAE-CNN; (d) BLS; (e) SVM; (f) KELM; (g) MLS-KELM

3) HOUSTON 2013 DATASET

TABLE VI.
CLASSIFICATION RESULTS OF INDIAN PINES DATASET

| Percentage of training set | Evaluation metrics | PCA-CNN | CAE-CNN | SVM | BLS | KELM | MLS-KELM |
|---|---|---|---|---|---|---|---|
| 10% | OA (%) | 93.25% | 89.76% | 96.31% | 90.23% | 97.23% | 97.32% |
|  | AA (%) | 94.23% | 91.36% | 96.72% | 90.56% | 97.52% | 97.60% |
|  | Kappa (%) | 92.70% | 88.98% | 96.01% | 89.44% | 97.00% | 97.11% |
|  | Tra-time(s) | 306.12 | 345.01 | 131.65 | 9.55 | 4.60 | 47.93 |
| 5% | OA (%) | 83.43% | 83.43% | 91.37% | 88.32% | 92.79% | 94.02% |
|  | AA (%) | 86.16% | 86.16% | 91.25% | 89.13% | 93.34% | 94.43% |
|  | Kappa (%) | 82.18% | 82.18% | 90.67% | 87.38% | 92.20% | 93.53% |
|  | Tra-time(s) | 221.00 | 219.32 | 85.21 | 8.59 | 1.74 | 28.55 |

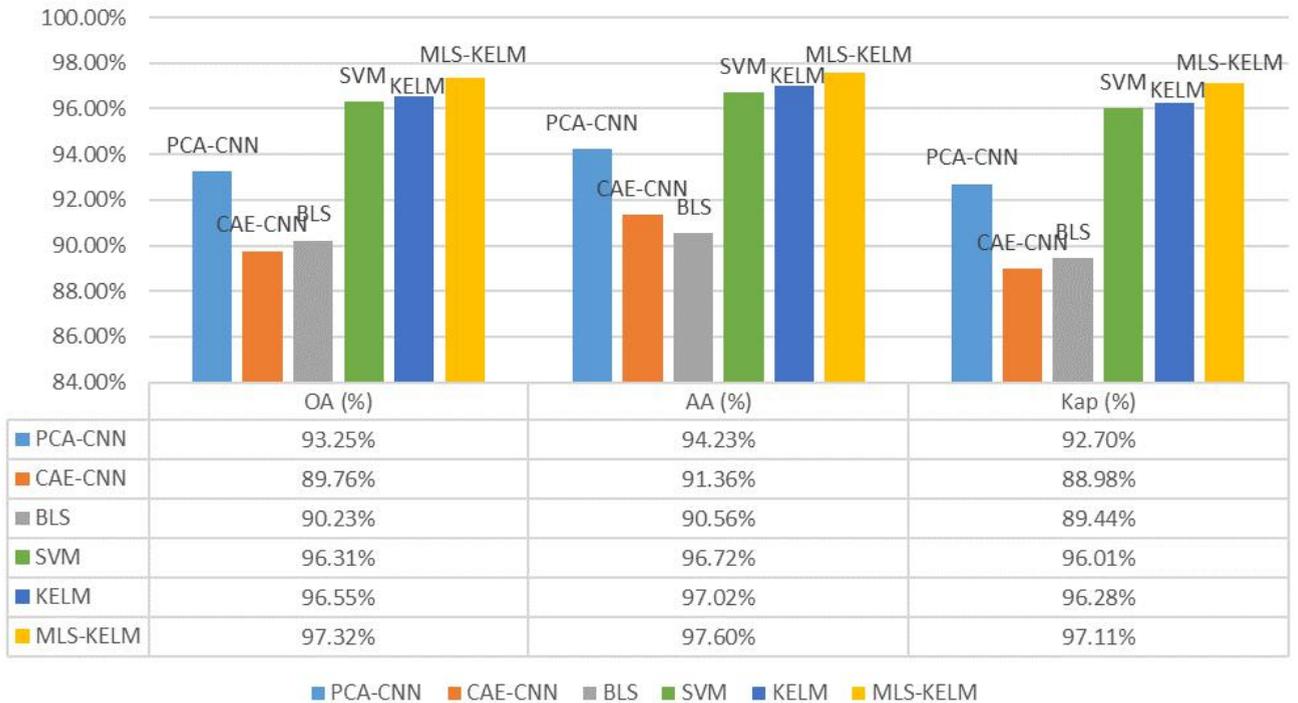

FIGURE.8 Classification comparison results on Houston 2013 dataset

The classification effects of each method in the Houston 2013 dataset are analyzed as follows: with a 10% training set, the OA, AA, and Kappa coefficients of KELM reach 97.23%, 97.52%, and 97.00%, respectively, which are significantly higher than those of PCA-CNN, CAE-CNN, BLS, and SVM, while the OA, AA, and Kappa coefficients of MLS-KELM reach 97.32%, 97.60%, and 97.11%, respectively, on which its OA, AA, and Kappa coefficients improved by 0.09%, 0.08%, and 0.11%, respectively. And the MLS-KELM method improves its OA, AA, and Kappa coefficients by 3.97%, 3.52%, and 4.28% on average with 10% training data relative to the popular PCA-CNN, CAE-CNN, BLS, SVM, and KELM methods. Significantly better than PCA-CNN, CAE-CNN, BLS, SVM, and KELM methods. With a 5% training set, MLS-KELM improves the OA, AA, and Kappa coefficients by 1.23%, 1.09%, and 1.33%, respectively, compared to the KELM method, and by 1.14%, 1.01%, and 1.22%, respectively, relative to the 10% training set. The experimental results show that MLS-KELM has high classification accuracy on Indian Pines and maintains a high classification performance in the case of small samples, indicating its strong robustness.

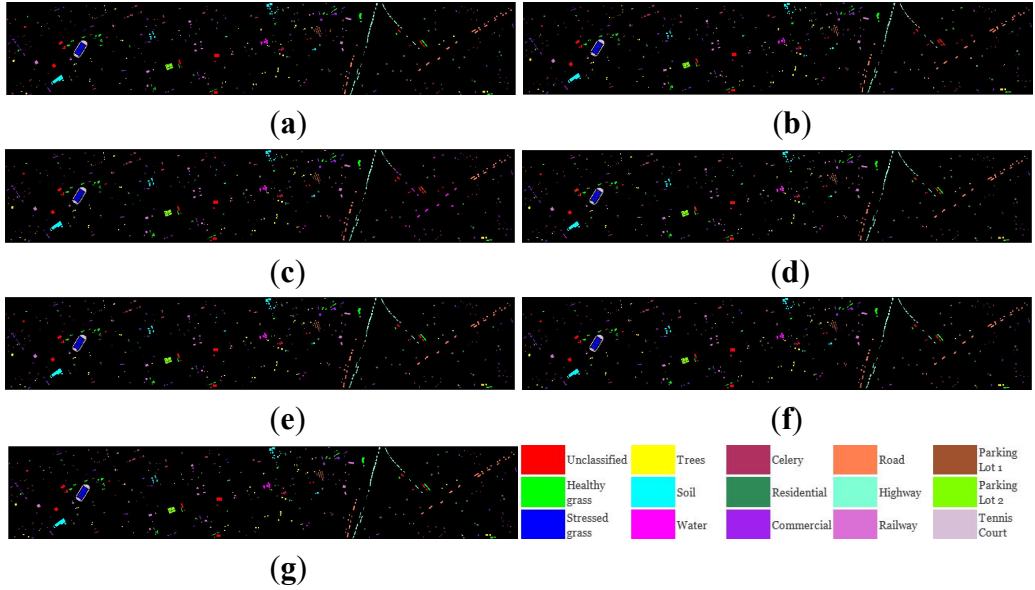

FIGURE.9 Classification effect of Houston 2013 dataset (10%). (a) Original graph; (b) PCA-CNN; (c) CAE-CNN; (d) BLS; (e) SVM; (f) KELM; (g) MLS-KELM.

In summary, these results show that SSA can effectively optimize KELM, making the MLS-KELM method have higher classification accuracy and generalization performance compared with the rest of the methods. And under each dataset, the improvement of MLS-KELM relative to KELM when taking a 5% training set is significantly higher than that when taking a 10% training set, which shows that MLS-KELM still maintains its high classification performance in the small sample case, indicating that MLS-KELM has strong robustness.

## V. APPLICATION IN ZY1-02D HYPERSPECTRAL DATA

To verify the universality and advancedness of the MLS-KELM, the hyperspectral remote sensing images taken by the ZY1-02D satellite are selected in this paper. Selected remote sensing images of a coastal area of Hainan Island (HI), China, which is a typical offshore ecosystem of Hainan Island.

The ZY1-02D hyperspectral data is a Class 1A product and can be downloaded from the Natural Resources Satellite Remote Sensing (NRSRSRS) China Cloud Service Platform (http://sasclouds.com/chinese/normal/). In this part, the images were pre-processed, including radiometric calibration, atmospheric correction, orthorectification correction, image cropping, etc. The bad bands affected by water vapor and clouds are removed (The bad bands of HI: 98-103,125-133,165-166). The images were finally labeled with six types of features, namely Sea, Lake, Shrubwood, Gravel, Colemanite, and Grey Leave, after MNF transformation, PPI extraction of pure image elements, n-dimensional visualization, and spectral angle mapping. The cropped field satellite image of Hainan Island (HI), China, is shown in Figure 10. The basic information and sample information of the HI dataset are shown in Table VII and Table VIII, respectively

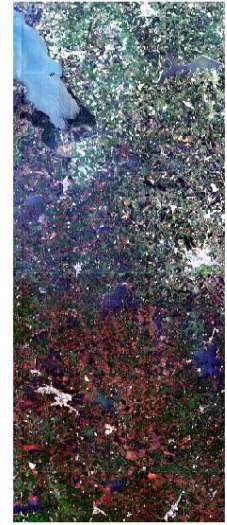

FIGURE.10 Satellite images of the HI dataset.

TABLE VII.
BASIC INFORMATION ABOUT THE HI DATASET.

| Data | HI data |
| --- | --- |
| Collection location | Hai Nan, China |
| Acquisition equipment | AHSI |
| Spectral coverage(μm) | 0.40~2.50 |
| Data size(pixel) | 1345×549 |
| Spatial resolution(m) | 30 |
| Number of bands | 166 |
| Number of bands after denoising | 149 |
| Sample size | 46537 |
| Number of categories | 6 |

TABLE VIII.
INFORMATION ABOUT THE HI DATASET.

| Category | Class name | Number of samples |
|---|---|---|
| 1 | Sea | 24904 |
| 2 | Lake | 10422 |
| 3 | Shrubwood | 9972 |
| 4 | Gravel | 169 |
| 5 | Colemanite | 388 |
| 6 | Grey Leave | 682 |

In this experiment, PCA-CNN, CAE-CNN, SVM, BLS, and KELM classification methods are still used as comparison method against the MLS-KELM method. 10% of the labeled features are divided into training sets for training the classification models. And 5% of the small sample training set is divided to validate the robustness of the method. The classification results for the HI dataset are shown in Table IX and Figure 11.

Analyzing the experimental results, we can see that the OA, AA, and Kappa coefficients of the MLS-KELM reach 96.31%, 94.59%, and 94.03%, respectively, in the case of 10% labeled samples, which is better than the PCA-CNN, CAE-CNN, SVM, BLS, and KELM classification models. The OA, AA, and Kappa coefficients are improved by 0.34%, 0.06%, and 0.57%, respectively, relative to KELM. With 5% labeled samples, the OA, AA, and Kappa coefficients of MLS-KELM reached 95.59%, 95.63%, and 92.82%, respectively, which were improved by 0.88%, 0.69%, and 1.05%, respectively, concerning the KELM. Experimental results show that MLS-KELM exhibits good classification results in ZY1-02D hyperspectral data. The improvement of MLS-KELM is greater when the labeled samples are reduced, which shows that MLS-KELM has good robustness.

The experimental results illustrate the advanced and universal applicability of the MLS-KELM and the great potential of China's ZY1-02D satellite for applications such as coastal monitoring.

TABLE IX.
CLASSIFICATION RESULTS FOR THE HI DATASET.

| Percentage of training set (%) | Evaluation metrics | PCA-CNN | CAE-CNN | SVM | BLS | KELM | MLS-KELM |
|---|---|---|---|---|---|---|---|
| 10% | OA (%) | 86.82% | 92.91% | 95.87% | 94.41% | 95.98% | 96.31% |
|  | AA (%) | 90.68% | 91.58% | 94.22% | 81.70% | 94.53% | 94.59% |
|  | Kappa (%) | 79.27% | 88.62% | 93.31% | 90.98% | 93.50% | 94.03% |
|  | Tra-time(s) | 220.19 | 211.2 | 81.75 | 10.37 | 10.73 | 26.41 |
| 5% | OA (%) | 88.33% | 91.11% | 94.34% | 87.40% | 94.76% | 95.59% |
|  | AA (%) | 89.13% | 90.89% | 93.97% | 81.05% | 94.97% | 95.63% |
|  | Kappa (%) | 85.56% | 88.72% | 92.33% | 80.06% | 91.86% | 92.82% |
|  | Tra-time(s) | 160.19 | 151.2 | 63.93 | 6.18 | 5.25 | 15.31 |

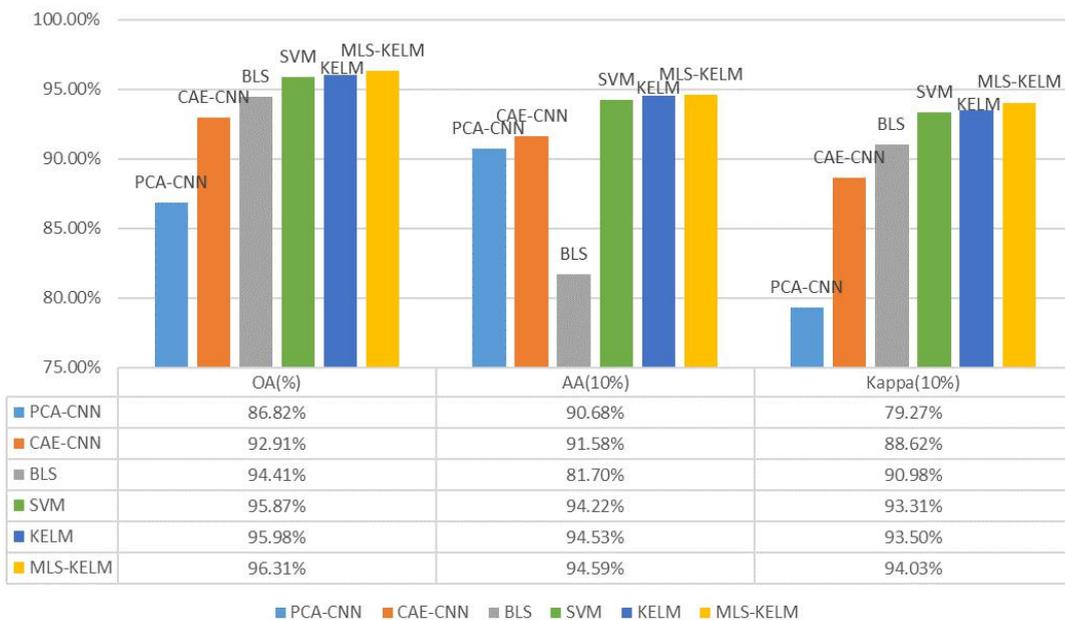

FIGURE.11 Classification comparison results on HI dataset.

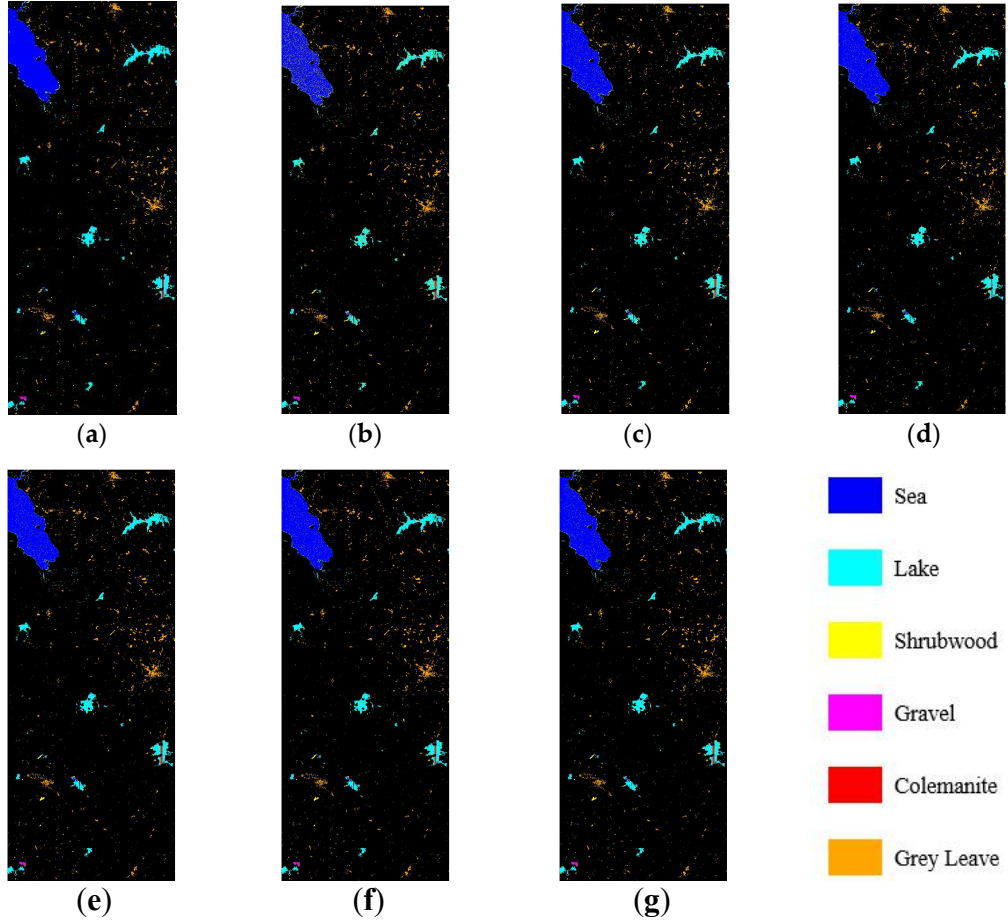

**FIGURE.12** Classification effect of HI dataset (10%). (a) Original graph; (b) PCA-CNN; (c) CAE-CNN; (d) BLS; (e) SVM; (f) KELM; (g) MLS-KELM.

## VI. DISCUSSION

In this study, the multiscale features of hyperspectral images are extracted and fused, and the swarm intelligence optimization algorithm is effectively combined with the popular classification algorithm to form an optimized image classifier. In this paper, we demonstrate that the MLS-KELM effectively solves the problem of low classification accuracy of hyperspectral images due to the presence of a large number of redundant bands and the Hugh phenomenon.

In terms of feature extraction of images, we have considered using popular feature extraction algorithms such as PCA, ICA, and LDA, but the above methods cannot resist the effect of texture noise well, which results in low discriminative spectral features. MSTV is a new noisy image feature extraction algorithm, which takes advantage of the feature that hyperspectral images have multi-scale features and uses the RTV model to set different parameters, multi-scale extraction of image spectral features, and noise reduction for bands of different dimensions, which greatly reduces the negative impact of texture noise. In this paper, we use the popular texture feature extraction operator-LBP, which has the advantages of grayscale invariance, rotation invariance, and fast speed. LBP has been applied to the texture descriptions of various images many times. And the features are fused by feature superposition, which can retain the original features well. The experiments in this paper prove that the feature extraction by MSTV combined with LBP is efficient.

This paper also explores the convergence and solution accuracy of various popular swarm intelligence optimization algorithms in processing hyperspectral data. The convergence graphs show that the GWO and WOA have stable iterative performance under each data set, and both have obtained better optimization results. However, they still have limitations relative to the SSA. The SSA, as a new type of swarm intelligence optimization with high convergence and excellent global search capability, has the best performance among the algorithms. The experiments prove that SSA has excellent advantages for processing hyperspectral data. Therefore, in our future work, we consider combining SSA with other popular algorithms to make more attempts.

KELM is an excellent classifier, but the kernel parameters and regularization coefficients need to be determined manually. Therefore, in this paper, a swarm intelligence optimization algorithm with high convergence and strong global search capability, SSA, is used to optimize KELM and determine the parameters efficiently and accurately. The spectral features of hyperspectral images are extracted by MSTV, the spatial features are extracted by LBP, and the fusion features of hyperspectral images are obtained by feature superposition. A high-precision multiscale fusion feature hyperspectral image classification method, MLS-KELM, is proposed, which effectively solves the problems of low accuracy of hyperspectral image classification and difficult processing of noisy images.

To verify the performance and universality of MLS-KELM, not only the popular Indian Pines, Houston 2013, and Pavia University datasets were selected, but also the practical applications of ZY1-02D hyperspectral data were performed, and MLS-KELM achieved good results for all of them. This shows that the ZY1-02D satellite has a strong potential for application. In this paper, we set up validation experiments for each dataset with small samples to verify whether the method can maintain its high classification accuracy despite the small number of labeled samples. The OA, AA, and Kappa coefficients of the MLS-KELM increased in the small sample experiments for the four datasets, indicating that the method has good robustness.

## VI. CONCLUSION

Since hyperspectral images are characterized by a large number of bands, narrow band spacing, large data redundancy, Hugh phenomenon, and the problems of " the same thing different spectrum" and " foreign body with the spectrum", it is difficult for traditional machine learning methods to effectively solve these problems. Therefore, this paper proposes a new hyperspectral classification method named MLS-KELM by combining MSTV, LBP, SSA, and KELM.

Two sets of experiments are set up in this paper. Experiment 1 is to compare the optimization effect and convergence performance of SSA with each popular swarm intelligence optimization method, and the results show that SSA still maintains its advantages of high convergence and strong global searchability in the field of hyperspectral classification. Therefore, in this paper, the optimal KELM classifier is derived by optimizing the kernel parameters and regularization coefficients with the high convergence and excellent global search ability of SSA. In Experiment 2, Indian Pines dataset, Houston dataset, and Pavia University dataset were selected, and two sets of experiments were set up with 10% and 5% training sets, respectively, to compare the MLS-KELM method with each popular classification method. In the Pavia University dataset, MLS-KELM achieves 99.28%, 99.54%, and 99.01% of OA, AA, and Kappa coefficients under the training set, respectively, and maintains its high classification accuracy in the rest of the dataset, indicating that MLS-KELM has better classification and generalization performance. Moreover, the MLS-KELM method improves the classification performance more in small samples, which indicates that the method has better robustness.

In this paper, a practical application of ZY1-02D hyperspectral data was carried out. The best results of MLS-KELM were also achieved in the practical application of ZY1-02D hyperspectral data. It shows that ZY1-02D has good potential for application.

Due to the lower operating efficiency of the MLS-KELM, in future work, we will consider how to improve its operational efficiency so that it can be better put into practical applications.